\documentclass[twocolumn]{svjour3}
\usepackage[]{microtype}
\usepackage{lineno,hyperref}
\usepackage{numberedblock}

\journalname{Soft Computing}

\usepackage{amsfonts,amsmath,amssymb}
\usepackage{url}
\usepackage{subfig}
\usepackage{multirow}
\usepackage{floatrow}

\usepackage{tikz}
\usepackage{csvsimple,booktabs,siunitx}
\usepackage[numbers]{natbib}

\usepackage[inline]{enumitem}

\newcommand{\bs}[1]{\boldsymbol{#1}}

\newcommand{\ie}{{i.e.}}
\newcommand{\eg}{{e.g.}}
\newcommand{\etal}{et al.}

\newcolumntype{L}[1]{>{\raggedright\arraybackslash}p{#1}}
\newcolumntype{C}[1]{>{\centering\arraybackslash}p{#1}}
\newcolumntype{R}[1]{>{\raggedleft\arraybackslash}p{#1}}

\begin{document}
\title{Deep Packet:  A Novel Approach For Encrypted Traffic Classification Using Deep Learning}

\author{Mohammad~Lotfollahi \and Mahdi~Jafari~Siavoshani \and Ramin~Shirali~Hossein~Zade  \and Mohammdsadegh~Saberian}

\institute{Mohammad~Lotfollahi \at Sharif University of Technology, Teharn, Iran
\and 
 Mahdi~Jafari~Siavoshani \at Sharif University of Technology, Teharn, Iran\\
\email{mjafari@sharif.edu} 
\and
Ramin~Shirali~Hossein~Zade \at Sharif University of Technology, Teharn, Iran
\and 
Mohammdsadegh~Saberian \at Sharif University of Technology, Teharn, Iran
}




\date{Received: date / Accepted: date}
\maketitle

\begin{abstract}
Internet traffic classification has become more important with rapid growth of current Internet network and online applications. There have been numerous studies on this topic which have led to many different approaches. Most of these approaches use predefined features extracted by an expert in order to classify network traffic. In contrast, in this study, we propose a \emph{deep learning}  based approach which integrates both feature extraction and classification phases into one system. Our proposed scheme, called ``Deep Packet,'' can handle both \emph{traffic characterization} in which the network traffic is categorized into major classes (\eg, FTP and P2P) and \emph{application identification} in which end-user applications (\eg, BitTorrent and Skype) identification is desired. Contrary to most of the current methods, Deep Packet can identify encrypted traffic and also distinguishes between VPN and non-VPN  network traffic. After an initial pre-processing phase on data, packets are fed into Deep Packet framework that embeds stacked autoencoder  and convolution neural network in order to classify network traffic. Deep packet with CNN as its  classification model achieved recall of $0.98$ in application identification task and $0.94$ in traffic categorization task. To the best of our knowledge, Deep Packet outperforms all of the proposed classification methods on UNB ISCX VPN-nonVPN dataset.
\keywords{Application Identification \and Traffic Characterization \and Deep Learning \and Convolutional Neural Networks \and Stacked Autoencoder \and Deep Packet.}
\end{abstract}


\section{Introduction} \label{sec:Introduction}
 
Network \emph{traffic classification} is an important task in modern communication networks \cite{bagui2017comparison}. Due to the rapid growth of high throughput traffic demands, to properly manage network resources, it is vital to recognize different types of applications utilizing network resources.
Consequently, accurate traffic classification has become one of the prerequisites for advanced network management tasks such as providing appropriate Quality-of-Service (QoS), anomaly detection, pricing, etc.
Traffic classification has attracted a lot of interests in both academia and industrial activities related to network management (\eg, see \cite{dainotti2012issues}, \cite{finsterbusch2014survey}, \cite{velan2015survey} and the references therein).

As an example of the importance of network traffic classification, one can think of the asymmetric architecture of today's network access links, which has been designed based on the assumption that clients download more than what they upload. However, the pervasiveness of symmetric-demand applications (such as peer-to-peer (P2P) applications, voice over IP (VoIP) and video call) has changed the clients' demands to deviate from the assumption mentioned earlier. Thus, to provide a satisfactory experience for the clients, other application-level knowledge is required to allocate adequate resources to such applications.

The emergence of new applications as well as interactions between various components on the Internet has dramatically increased the complexity and diversity of this network which makes the traffic classification a difficult problem per se. In the following, we discuss in details some of the most critical challenges of network traffic classification.


First, the increasing demand for user's privacy and data encryption has tremendously raised the amount of encrypted traffic in today's Internet \cite{velan2015survey}.
Encryption procedure turns the original data into a pseudo-random-like format to make it hard to decrypt. In return, it causes the encrypted data scarcely contain any discriminative patterns to identify network traffic.
Therefore, accurate classification of encrypted traffic has become a challenge in modern networks \cite{dainotti2012issues}.

It is also worth mentioning that many of the proposed network traffic classification approaches, such as payload inspection as well as machine learning and statistical methods, require patterns or features to be extracted by experts. This process is prone to error, time-consuming and costly.

Finally, many of the Internet service providers (ISPs) block P2P file sharing applications because of their high bandwidth consumption and copyright issues \cite{lv2014deepflow}. These applications use protocol embedding techniques to bypass traffic control systems \cite{alshammari2011can}. 
The identification of this kind of applications is one of the most challenging task in network traffic classification.

There have been abundant studies on the network traffic classification subject, \eg,  \cite{KohoutNetworkTrafficFingerprinting2018,PereraComparisonSupervisedMachine2017,gil2016characterization,moore2005toward}. However, most of them have focused on classifying a protocol family, also known as \emph{traffic characterization} (\eg, streaming, chat, P2P, etc.), instead of identifying a single application, which is known as \emph{application identification} (\eg, Spotify, Hangouts, BitTorrent, etc.) \cite{khalife2014multilevel}. 
In contrast, this work proposes a method, \ie, Deep Packet, based on the ideas recently developed in the machine learning community, namely, deep learning, \cite{Bengio:2009_NowPublishers,lecun2015deep_Nature}, to both characterize and identify the network traffic. The benefits of our proposed method which make it superior to other classification schemes are stated as follows:

\begin{itemize}
\item In Deep Packet, there is no need for an expert to extract features related to network traffic. In the light of this approach, the cumbersome step of finding and extracting distinguishing features has been omitted. 

\item Deep Packet can identify traffic at both granular levels (application identification and traffic characterization) with state-of-the-art results compared to the other works conducted on similar dataset \cite{gil2016characterization,yamansavascilar2017application}.

\item Deep Packet can accurately classify one of the hardest class of applications, known to be P2P \cite{khalife2014multilevel}. This kind of applications routinely uses advanced port obfuscation techniques, embedding their information in well-known protocols' packets and using random ports to circumvent ISPs' controlling processes.
\end{itemize}

The rest of paper is organized as follows. In Section~\ref{sec:relatedWorks}, we review some of the most important and recent studies on traffic classification. In Section~\ref{sec:Background}, we present the essential background on deep learning which is necessary to our work. Section~\ref{sec:Methodology} presents our proposed method, \ie, Deep Packet. The results of the proposed scheme on network application identification and traffic characterization tasks are described in Section~\ref{sec:results}. In  Section~\ref{sec:discus}, we provide further discussion on experimental results. Finally, we conclude the paper in Section~\ref{sec:conc}.

\section{Related Works}\label{sec:relatedWorks}
In this section, we provide an overview of the most important network traffic classification methods. In particular, we can categorize these approaches into three main categories as follows: 
\begin{enumerate*}[label={(\Roman*)}]	
  \item port-based,
  \item payload inspection, and
  \item statistical and machine learning.
\end{enumerate*}
Here is a brief review of the most important and recent studies regarding each of the approaches mentioned above.

\textbf{Port-based approach:}
Traffic classification via port number is the oldest and the most well-known method for this task \cite{dainotti2012issues}. Port-based classifiers use the information in the TCP/UDP headers of the packets to extract the port number which is assumed to be associated with a particular application. After the extraction of the port number, it is compared with the assigned IANA TCP/UDP port numbers for traffic classification. The extraction is an easy procedure, and port numbers will not be affected by encryption schemes. Because of the fast extraction process, this method is often used in firewall and access control list (ACL) \cite{qi2009packet}. Port-based classification is known to be among the simplest and fastest method for network traffic identification. However, the pervasiveness of port obfuscation, network address translation (NAT), port forwarding, protocol embedding and random ports assignments have significantly reduced the accuracy of this approach. According to \cite{moore2005toward,madhukar2006longitudinal} only $30\%$ to $70\%$  of the current Internet traffic can be classified using port-based classification methods. For these reasons, more complex traffic classification methods are needed to classify modern network traffic.

\textbf{Payload Inspection Techniques:} These techniques are based on the analysis of information available in the application layer payload of packets \cite{khalife2014multilevel}. Most of the payload inspection methods, also known as deep packet inspection (DPI), use predefined patterns like regular expressions as signatures for each protocol (\eg, see \cite{yeganeh2012cute,Sen:2004:ASI:988672.988742}). The derived patterns are then used to distinguish protocols form each other. The need for updating patterns whenever a new protocol is released, and user privacy issues are among the most important drawbacks of this approach. Sherry et al. proposed a new DPI system that can inspect encrypted payload without decryption, thus solved the user privacy issue, but it can only process HTTP Secure (HTTPS) traffic \cite{sherry2015blindbox}.


\textbf{Statistical and machine learning approach:}\\
Some of these methods, mainly known as statistical methods, have a biased assumption that the underlying traffic for each application has some statistical features which are almost unique to each application. Each statistical method uses its functions and statistics. Crotti et al., \cite{crotti2007traffic} proposed protocol fingerprints based on the probability density function (PDF) of packets inter-arrival time and normalized thresholds. They achieved up to $91\%$ accuracy for a group of protocols such as HTTP, Post Office Protocol 3 (POP3) and Simple Mail Transfer Protocol (SMTP). In a similar work, Wang et al., \cite{wang2010optimised} have considered PDF of the packet size. Their scheme was able to identify a broader range of protocols including file transfer protocol (FTP), Internet Message Access Protocol (IMAP), SSH, and TELNET with accuracy up to $87\%$.

A vast number of machine learning approaches have been published to classify traffic. Auld et al. proposed a  Bayesian neural network that was trained to classify most well-known P2P protocols including Kazaa, BitTorrent, GnuTella, and achieved $99\%$ accuracy \cite{auld2007bayesian}. Moore et al. achieved $96\%$ of accuracy on the same set of applications using a Naive Bayes classifier and a kernel density estimator \cite{moore2005internet}. Artificial neural network (ANN) approaches were proposed for traffic identification (e.g., see \cite{sun2010traffic} and \cite{ting2010network}). Moreover, it was shown in \cite{ting2010network} that the ANN approach can outperform Naive Bayes methods.  Two of the most important papers that have been published on ``ISCX VPN-nonVPN" traffic dataset are based on machine learning methods. Gil et al.  used time-related features such as the duration of the flow, flow bytes per second, forward and backward inter-arrival time, etc. to characterize the network traffic using k-nearest neighbor (k-NN) and C4.5 decision tree algorithms \cite{gil2016characterization}. They achieved approximately $92\%$  recall, characterizing six major classes of traffic including Web browsing, email, chat, streaming, file transfer and VoIP using the C4.5 algorithm. They also achieved approximately $88\%$ recall using the C4.5 algorithm on the same dataset which is tunneled through VPN. Yamansavascilar et al. manually selected $111$  flow features described in \cite{moore2013discriminators} and achieved  $94\%$ of accuracy  for $14$ class of applications using k-NN algorithm \cite{yamansavascilar2017application}.  
The main drawback of all these approaches is that the feature extraction and feature selection phases are essentially done with the assistance of an expert. Hence, it makes these approaches time-consuming, expensive and prone to human mistakes. Moreover, note that for the case of using k-NN classifiers, as suggested by  \cite{yamansavascilar2017application}, it is known that, when used for prediction, the execution time of this algorithms  is a major concern.
        
To the best of our knowledge, prior to our work, only one study based on deep learning ideas has been reported by Wangc \cite{wang2015applications}. They used stacked autoencoders (SAE) to classify some network traffic for a large family of protocols like HTTP, SMTP, etc. However, in their technical report, they did not mention the dataset they used. Moreover, the methodology of their scheme, the details of their implementation, and the proper report of their result is missing.

\section{Background on Deep Neural Networks}\label{sec:Background}

Neural networks (NNs) are computing systems made up of some simple, highly interconnected processing elements, which process information by their dynamic state response to external inputs \cite{Caudill:1987:NNP:38292.38295}. In practice, these networks are typically constructed from a vast number of building blocks called \emph{neuron} where they are connected via some links to each other. These links are called connections, and to each of them, a weight value is associated.
During the training procedure, the NN is fed with a large number of data samples. The widely used learning algorithm to train such networks (called \emph{backpropagation}) adjusts the weights to achieve the desired output from the NN. The deep learning framework can be considered as a particular kind of NNs with many (hidden) layers. Nowadays, with the rapid growth of computational power and the availability of graphical processing units (GPUs), training deep NNs have become more plausible. Therefore, the researchers from different scientific fields consider using deep learning framework in their respective area of research, \eg, see \cite{hinton2012deep,simonyan2014very,socher2013recursive}.
In the following, we will briefly review two of the most important deep neural networks that have been used in our proposed scheme for network traffic classification, namely, autoencoders and convolutional neural networks.


\subsection{Autoencoder}\label{subsec:ae}
\begin{figure}[!htb] 
 	\centering
	  \includegraphics[width = 0.75\linewidth]{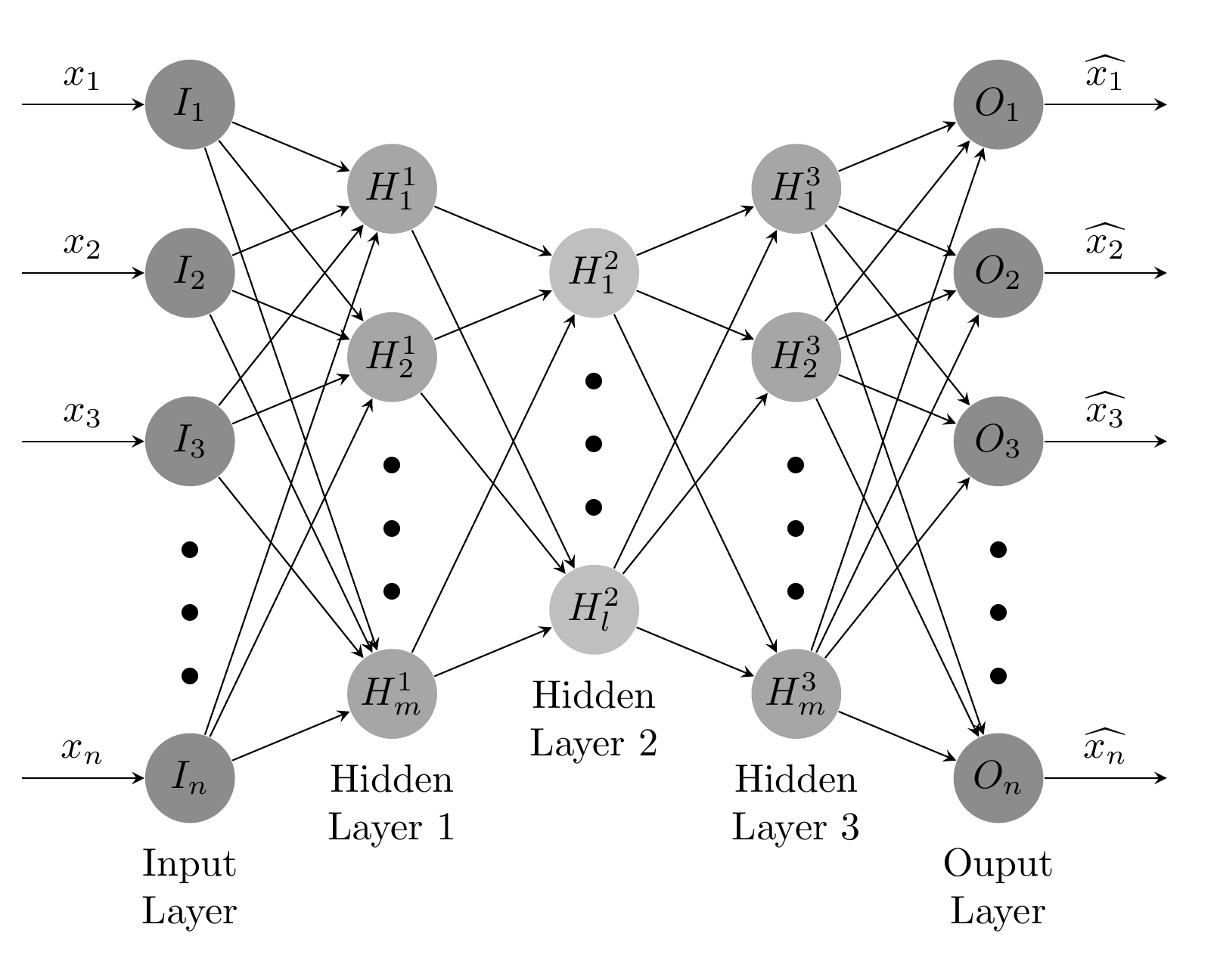}
  \caption{The general structure of an autoencoder.}
  \label{fig:boat1}
\end{figure}

An autoencoder NN is an unsupervised learning framework that uses backpropagation algorithm to reconstruct the input at the output while minimizing the reconstruction error (\ie, according to some criteria). Consider a training set $\{x^1,x^2,\ldots,x^n\}$ where for each training data we have $x^i \in \mathbb{R}^n$. The autoencoder objective is defined to be $y^i = x^i$ for $i\in\{1,2,\ldots,n\}$, \ie, the output of the network will be equal to its input. Considering this objective function, the autoencoder tries to learn a compressed representation of the dataset, \ie, it approximately learns the identity function $F_{\bs{W},\bs{b}}(x)\simeq x$, where $\bs{W}$ and $\bs{b}$ are the whole network weights and biases vectors. General form of an autoencoder's loss function is shown in \eqref{eq:auto}, as follows
\begin{equation}
\mathcal{L}(\bs{W},\bs{b}) = \left\| x-F_{\bs{W},\bs{b}}(x) \right\|^2.
\label{eq:auto}
\end{equation}
The autoencoder is mainly used as an unsupervised technique for automatic feature extraction. More precisely, the output of the encoder part is considered as a high-level set of discriminative features for the classification task. Fig.~\ref{fig:boat1} shows a typical autoencoder with $n$  inputs and outputs. 

\begin{figure}[!htb] 
\centering
  \includegraphics[width=.75\linewidth]{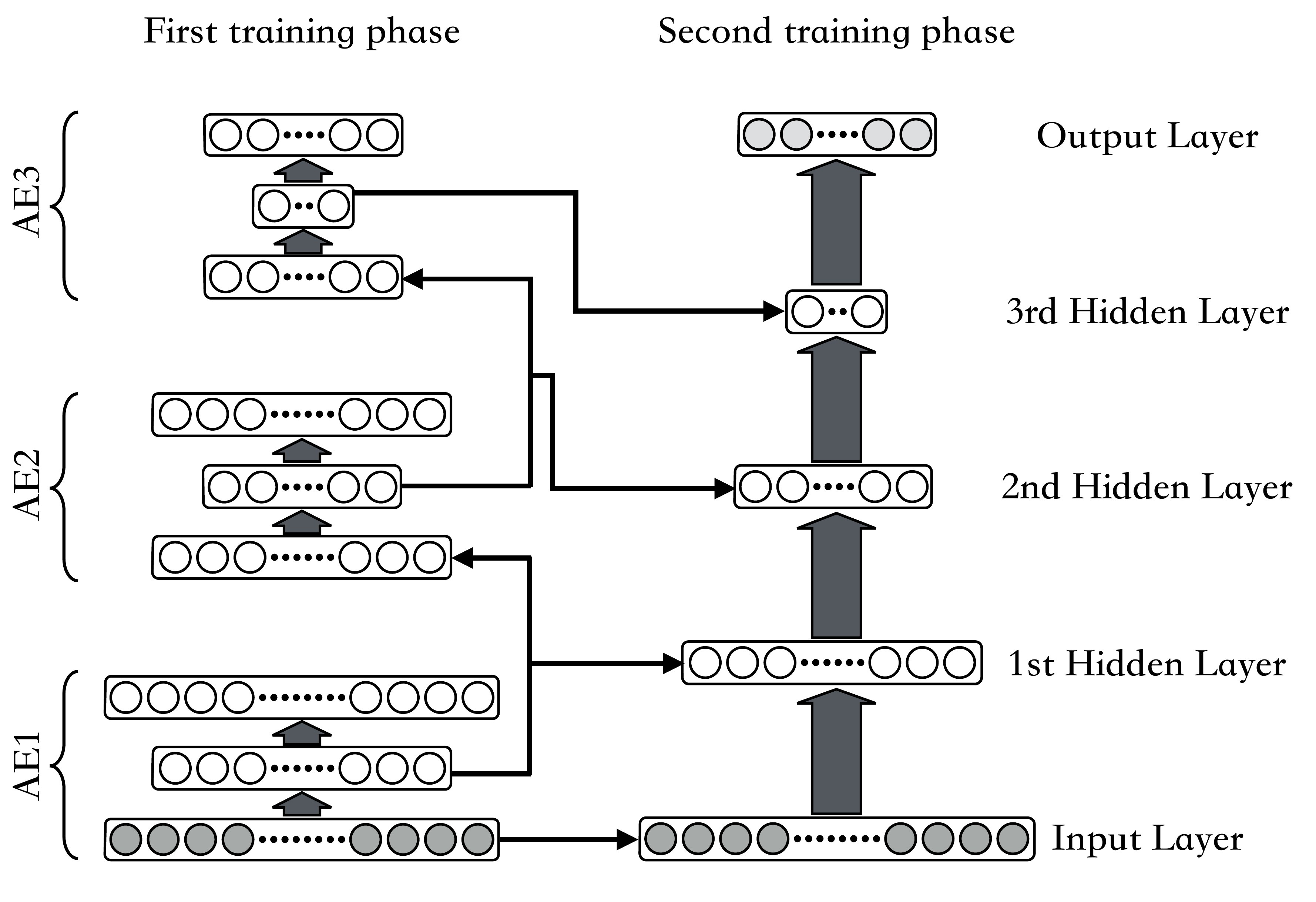}
  \caption{Greedy layer-wise approach for training an stacked autoendcoder.}
  \label{fig:sae}
 \end{figure}

In practice, to obtain a better performance, a more complex architecture and training procedure, called stacked autoencoder (SAE), is proposed \cite{vincent2008extracting}. This scheme suggests to stack up several autoencoders in a manner that output of each one is the input of the successive layer which itself is an autoencoder. The training procedure of a stacked autoencoder is done in a greedy layer-wise fashion \cite{bengio2007greedy}.
First, this method trains each layer of the network while freezing the weights of other layers. After training all the layers, to have more accurate results, fine-tuning is applied to the whole NN. At the fine-tuning phase, the backpropagation algorithm is used to adjust all layers' weights. 
Moreover, for the classification task, an extra softmax layer can be applied to the final layer. 
Fig.~\ref{fig:sae}, depicts the training procedure of a stacked autoencoder.

\subsection{Convolutional Neural Network}
The convolutional neural networks (CNN) are another types of deep learning models in which feature extraction from the input data is done using layers comprised of convolutional operations. The construction of convolutional networks is inspired by the visual structure of living organisms \cite{hubel1968receptive}. 
Basic building block underneath a CNN is a convolutional layer described as follows.
Consider a convolutional layer with $ N \times N$ square neuron layer as input and a filter $\omega$ of size $m\times m$. The output of this layer $z^l$, is of size $(N-m+1) \times (N-m+1)$ and is computed as follows
\begin{equation}\label{eq:conv_exp}
z_{ij}^\ell =f\left( \sum_{a=0}^{m-1} \sum_{b=0}^{m-1} \omega_{ab} z_{(i+a)(j+b)}^{\ell - 1} \right).
\end{equation}
As it is demonstrated in \eqref{eq:conv_exp},
usually a non-linear function $f$ such as rectified linear unit (ReLU) is applied to the convolution output to learn more complex features from the data. In some applications, a pooling layer is also applied. The main motivation of employing a pooling layer is to aggregate multiple low-level features in a neighborhood to obtain local invariance. Moreover, it helps to reduce the computation cost of the network in train and test phase.

CNNs have been successfully applied to different fields including natural language processing \cite{dos2014deep}, computational biology \cite{alipanahi2015predicting}, and machine vision \cite{simonyan2014very}. One of the most interesting applications of CNNs is in face recognition \cite{Lee:2009:CDB:1553374.1553453}, where consecutive convolutional layers are used to extract features from each image. It is observed that the extracted features in shallow layers are simple concepts like edges and curves. On the contrary, features in deeper layers of networks are more abstract than the ones in shallower layers \cite{DBLP:journals/corr/YosinskiCNFL15}. 
However, it is worth mentioning that visualizing the extracted features in the middle layers of a network does not always lead to meaningful concepts like what has been observed in the face recognition task. For example in one-dimensional CNN (1D-CNN) which we use to classify network traffic, the feature vectors extracted in shallow layers are just some real numbers which make no sense at all for a human observer. 

We believe 1D-CNNs are an ideal choice for the network traffic classification task. This is true since 1D-CNNs can capture spatial dependencies between adjacent bytes in network packets that leads to find discriminative patterns for every class of protocols/applications, and consequently, an accurate classification of the traffic. Our classification results confirm this claim and prove that CNNs performs very well in feature extraction of network traffic data.

\section{Methodology}
\label{sec:Methodology}

\begin{figure}[!htb] 
\centering
  \includegraphics[width=1\linewidth]{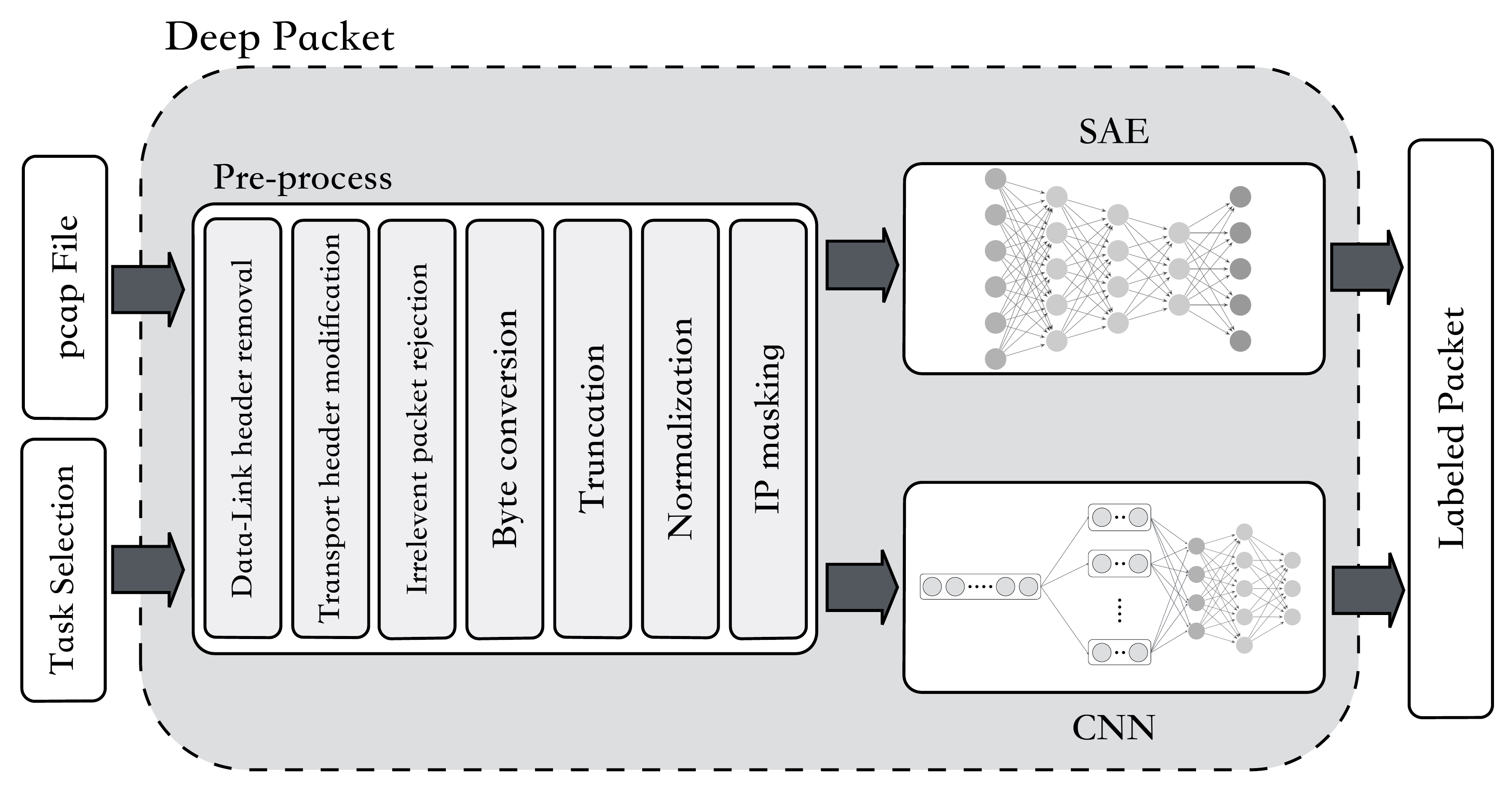}
  \caption{General illustration of Deep Packet toolkit.
}
  
  \label{fig:system}
 \end{figure}

In this work, we develop a framework, called Deep Packet, that comprises of two deep learning methods, namely, convolutional NN and stacked autoencoder NN, for both ``application identification'' and ``traffic characterization'' tasks. Before training the NNs, we have to prepare the network traffic data so that it can be fed into NNs properly. To this end, we perform a pre-processing phase on the dataset. Fig.~\ref{fig:system} demonstrates the general structure of Deep Packet. At the test phase, a pre-trained neural network corresponding to the type of classification, application identification or traffic characterization, is used to predict the class of traffic the packet belongs to. The dataset, implementation and design details of the pre-processing phase and the architecture of proposed NNs will be explained in the following.

\subsection{Dataset}\label{subsec:dataset}
For this work, we use ``ISCX VPN-nonVPN" traffic dataset, that consists of captured traffic of different applications in pcap format files \cite{gil2016characterization}. 
In this dataset, the captured packets are separated into different pcap files labeled according to the application produced the packets (\eg, Skype, and Hangouts, etc.) and the particular activity the application was engaged during the capture session (\eg, voice call, chat, file transfer, or video call). For more details on the captured traffic and the traffic generation process, refer to \cite{gil2016characterization}.

The dataset also contains packets captured over Virtual Private Network (VPN) sessions. A VPN 
 is a private overlay network among distributed sites which operates by tunneling traffic over public communication networks (\eg, the Internet). Tunneling IP packets, guaranteeing secure remote access to servers and services, is the most prominent aspect of VPNs \cite{chowdhury2010survey}. Similar to regular (non-VPN) traffic, VPN traffic is captured for different applications, such as Skype, while performing different activities, like voice call, video call, and chat. 

Furthermore, this dataset contains captured traffic of Tor software. This traffic is presumably generated while using Tor browser and it has labels such as Twitter, Google, Facebook, etc. Tor is a free, open source software developed for anonymous communications. Tor forwards users’ traffic through its own free, worldwide, overlay network which consists of volunteer-operated servers. Tor was proposed to protect users against Internet surveillance known as ``traffic analysis.'' To create a private network pathway, Tor builds a circuit of encrypted connections through relays on the network in a way that no individual relay ever knows the complete path that a data packet has taken \cite{dingledine2004tor}. Finally, Tor uses complex port obfuscation algorithm to improve privacy and anonymity.

\subsection{Pre-processing}\label{subsec:pre-proc}

The ``ISCX VPN-nonVPN" dataset is captured at the data-link layer. Thus it includes the Ethernet header. The data-link header contains information regarding the physical link, such as Media Access Control (MAC) address, which is essential for forwarding the frames in the network, but it is uninformative for either the application identification or traffic characterization tasks. Hence, in the pre-processing phase, the Ethernet header is removed first. Transport layer segments, specifically Transmission Control Protocol (TCP) or User Datagram Protocol (UDP), vary in header length. The former typically bears a header of $20$ bytes length while the latter has an $8$ bytes header. To make the transport layer segments uniform, we inject zeros to the end of UDP segment's headers to make them equal length with TCP headers. The packets are then transformed from bits to bytes which helps to reduce the input size of the NNs. 

Since the dataset is captured in a real-world emulation, it contains some irrelevant packets which are not of our interest and should be discarded. In particular, the dataset includes some TCP segments with either SYN, ACK, or FIN flags set to one and containing no payload. These segments are needed for three-way handshaking procedure while establishing a connection or finishing one, but they carry no information regarding the application generated them, thus can be discarded safely. Furthermore, there are some Domain Name Service (DNS) segments in the dataset. These segments are used for hostname resolution, namely, translating URLs to IP addresses. These segments are not relevant to either application identification or traffic characterization, hence can be omitted from the dataset.

\begin{figure}[!htb] 
\centering
  \includegraphics[width=\linewidth]{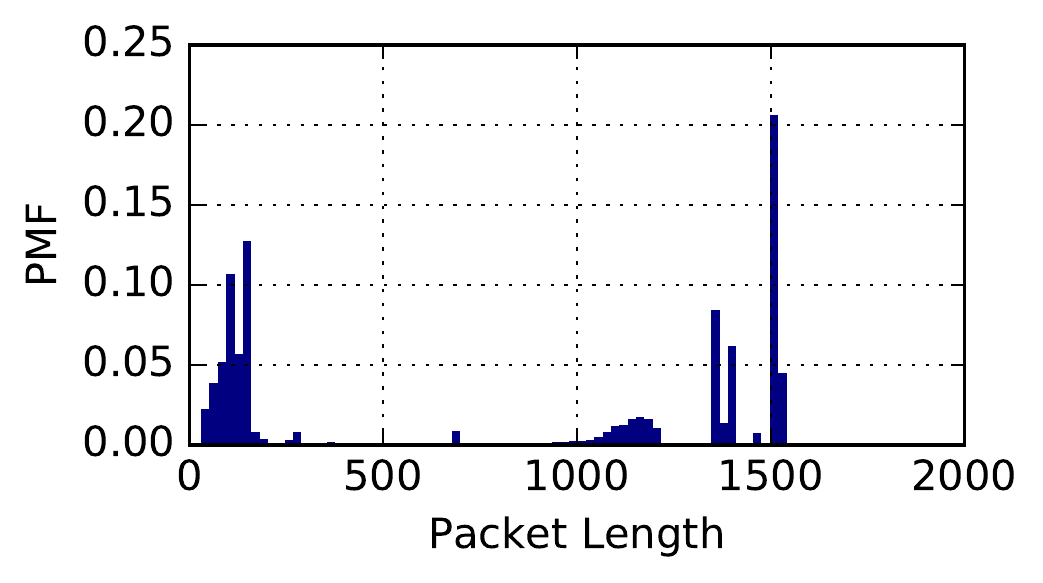}
  \caption{Empirical probability mass function of the packet length in ISCX VPN-nonVPN traffic dataset.}
  \label{fig:histogram}
 \end{figure}
 
Fig.~\ref{fig:histogram} illustrates the histogram (empirical distribution) of packet length for the dataset. As the histogram shows, packet length varies a lot through the dataset, while employing NNs necessitates using a fixed-size input. Hence, truncation at a fixed length or zero-padding are required inevitably. To find the fixed length for truncation, we inspected the packets length's statistics. Our investigation revealed that approximately $96\%$ of packets have a payload length of less than $1480$ bytes. This observation is not far from our expectation, as most of the computer networks are constrained by Maximum Transmission Unit (MTU) size of 1500 bytes. Hence, we keep the IP header and the first $1480$ bytes of each IP packet which results in a $1500$ bytes vector as the input for our proposed NNs. Packets with IP payload less than $1480$ bytes are zero-padded at the end. To obtain a better performance, all the packet bytes are divided by $255$, the maximum value for a byte, so that all the input values are in the range $[0,1]$.
 
Furthermore, since there is the possibility that the NN attempts to learn classifying the packets using their IP addresses, as the dataset is captured using a limited number of hosts and servers, we decided to prevent this over-fitting by masking the IP addresses in the IP header. In this matter, we assure that the NN is not using irrelevant features to perform classification.
All of the pre-processing steps mentioned above take place when the user loads a pcap file into Deep Packet toolkit.


\subsubsection{Labeling Dataset}\label{subsec:label}
As mentioned before in Section~\ref{subsec:dataset}, the dataset's pcap files are labeled according to the applications and activities they were engaged in. However, for application identification and traffic characterization tasks, we need to redefine the labels, concerning each task. For application identification, all pcap files labeled as a particular application which were collected during a nonVPN session,
are aggregated into a single file. This leads to 17 distinct labels shown in Table~\ref{table:classes}a. Also for traffic characterization, we aggregated the captured traffic of different applications involved in the same activity, taking into account the VPN or non-VPN condition, into a single pcap file. This leads to a 12-classes dataset, as shown in Table~\ref{table:classes}b. By observing Tables~\ref{table:classes}, one would instantly notice that the dataset is significantly imbalanced and the number of samples varies remarkably among different classes. It is known that such an imbalance in the training data leads to a reduced classification performance. Sampling is a simple yet powerful technique to overcome this problem \cite{longadge2013class}. Hence, to train the proposed NNs, using the under-sampling method, we randomly remove the major classes' samples (classes having more samples) until the classes are relatively balanced.

\begin{table}[!htb]
    \centering
    
    \scriptsize
    \caption[caption]{Number of samples in each class for (a) Application identification, and (b) Traffic characterization.}
    \label{table:classes}
    \subfloat[][]{
	\begin{tabular}{| c|c |}
	\hline 
Application & Size \\
    \hline
AIM chat & 5K \\ 
Email & 28K \\
Facebook & 2502K \\
FTPS & 7872K \\ 
Gmail &  12K \\
Hangouts & 3766K \\ 
ICQ & 7K \\ 
Netflix & 299K \\
SCP & 448K \\ 
SFTP & 418K \\ 
Skype & 2872K \\ 
Spotify & 40K \\ 
Torrent & 70K \\
Tor & 202K \\ 
Voipbuster & 842K \\ 
Vimeo & 146K \\ 
YouTube & 251K \\
\hline
	\end{tabular}
	\label{table:app}
}
\subfloat[][]{
\renewcommand{\arraystretch}{1.37}
\begin{tabular}{|c|c|}
	\hline
	Class Name & Size \\
	\hline
	Chat& 82K\\ 
	Email& 28K \\ 
	File Transfer&210K \\ 
	Streaming& 1139K\\ 
	Torrent&70K \\ 
	VoIP&5120K \\ 
	VPN: Chat&50K \\ 
	VPN: File Transfer& 251K \\ 
	VPN: Email&13K \\ 
	VPN: Streaming& 479K \\ 
	VPN: Torrent&269K \\ 
	VPN: VoIP&753K \\ 
	\hline
\end{tabular}
	\label{table:char}}
\end{table}

\subsection{Architectures}

The proposed SAE architecture consists of five fully-connected layers, stacked on top of each other which made up of $400$, $300$, $200$, $100$, and $50$ neurons, respectively. To prevent the over-fitting problem, after each layer the dropout technique with $0.05$ dropout rate is employed. In this technique, during the training phase, some of the neurons are set to zero randomly. Hence, at each iteration, there is a random set of active neurons. For the application identification and traffic characterization tasks, at the final layer of the proposed SAE, a softmax classifier with $17$ and $12$ neurons are added, respectively. 

\begin{figure}
\centering
\includegraphics[width=\linewidth]{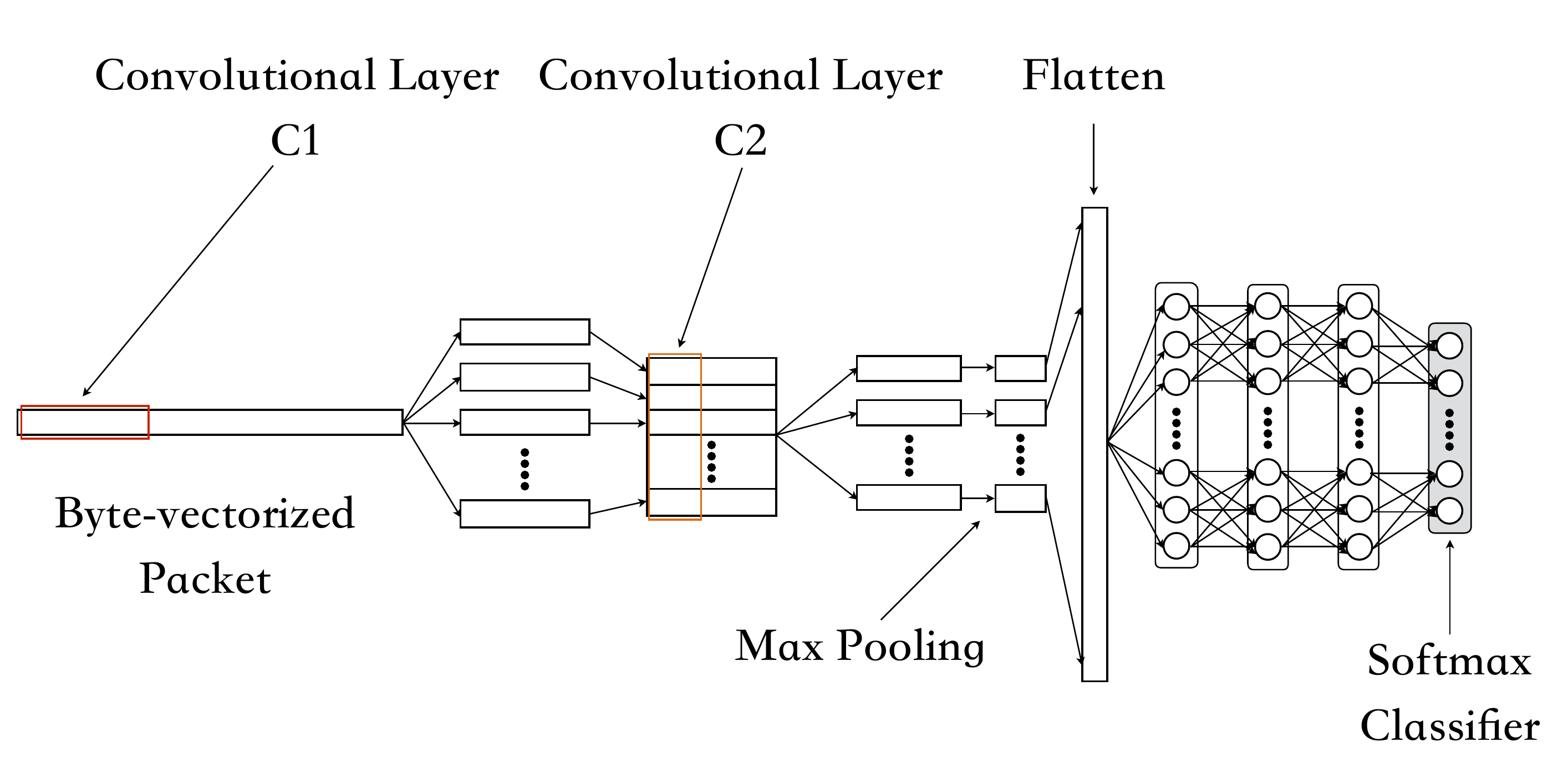}\label{fig:convolution-arch}
\caption{A minimal illustration of the proposed one-dimensional CNN architecture.}%
\label{fig:archs}
\end{figure}

A minimal illustration of the second proposed scheme, based on one-dimensional CNN, is depicted in Fig.~\ref{fig:archs}. We used a grid search on a subspace of the hyper-parameters space to select the ones which results in the best performance. This procedure is discussed in detail in the Section \ref{sec:results}. The model consists of two consecutive convolutional layers, followed by a pooling layer. Then the two-dimensional tensor is squashed into a one-dimensional vector, and fed into a three-layered network of fully connected neurons which also employ dropout technique to avoid over-fitting. Finally, a softmax classifier is applied for the classification task, similar to SAE architecture. The best values found for the hyper-parameters are shown in Table~\ref{table:params}.

\begin{table}[!htb]
    \centering
    
    \scriptsize
    \caption[caption]{Selected hyper-parameters for the CNNs.}
    \label{table:params}

    \makebox[0.5\linewidth]{
	\begin{tabular}{|c|c|c|c|c|c|c|}
	\hline 
\multirow{2}{*}{Task} &\multicolumn{3}{c|}{C1 Filter} & \multicolumn{3}{c|}{C2 Filter} \\
   \cline{2-7}
& Size & Number & Stride & Size & Number & Stride\\
\hline
App. Idn. & {4}& {200}& {3}& {5}& {200}& {1} \\ 
\hline
Traffic Char. & {5}& {200}& {3}&{4}&200 & {3} \\ 
\hline
	\end{tabular}
	
}

\end{table}


%


\section{Experimental Results}\label{sec:results}
To implement our proposed NNs, we have used Keras library \cite{chollet2015keras}, with Tensorflow \cite{tensorflow2015-whitepaper} as its backend. Each of the proposed models was trained and evaluated against the independent test set that was extracted from the dataset. We randomly split the dataset into three separate sets: the first one which includes $64\%$ of samples is used for training and adjusting weights and biases; the second part containing $16\%$ of samples is used for validation during the training phase; and finally the third set made up of $20\%$ of data points are used for testing the model. Additionally, to avoid the over-fitting problem, we have used \emph{early stopping} technique \cite{prechelt1998early}. This technique stops the training procedure, once the value of loss function on the validation set remains almost unchanged for several epochs, thus prevents the network to over-fit on the training data. To speed up the learning phase, we also used \emph{Batch Normalization} technique in our models \cite{ioffe2015batch}.

For training SAE, first each layer was trained in a greedy layer-wise fashion using \emph{Adam} optimizer \cite{kingma2014adam} and \emph{mean squared error} as the loss function for $200$ epochs, as described in Section~\ref{subsec:ae}. Next, in the fine-tuning phase, the whole network was trained for another $200$ epochs using the \emph{categorical cross entropy} loss function.  Also, for implementing the proposed one dimensional CNN, the \emph{categorical cross entropy} and \emph{Adam} were used as loss function and optimizer, respectively, and in this case, the network was trained for $300$ epochs.
Finally, it is worth mentioning that in both NNs, all layers employs Rectified Linear Unit (ReLU) as the activation function, except for the final softmax classifier layer.

To evaluate the performance of Deep Packet, we have used Recall (Rc), Precision (Pr) and $F_1$ Score (\ie, $F_1$) metrics. The above metrics are described mathematically as follows:
 \begin{equation}\label{eq:metrics}
 \small
\textrm{Rc} =  \frac{\textrm{TP}}{\textrm{TP} + \textrm{FN}}, \quad \textrm{Pr} = \frac{\textrm{TP}}{\textrm{TP} + \textrm{FP}}, \quad  F_1 =  \frac{2\cdot\textrm{Rc} \cdot \textrm{Pr} }{\textrm{Rc} + \textrm{Pr}}.
\end{equation}
\normalsize
Here, TP, FP and FN stands for true positive, false positive and false negative, respectively.


\begin{figure}[!htb]
\centering
\subfloat[][Aplication Identification]{\includegraphics[width=0.9\linewidth,page=1]{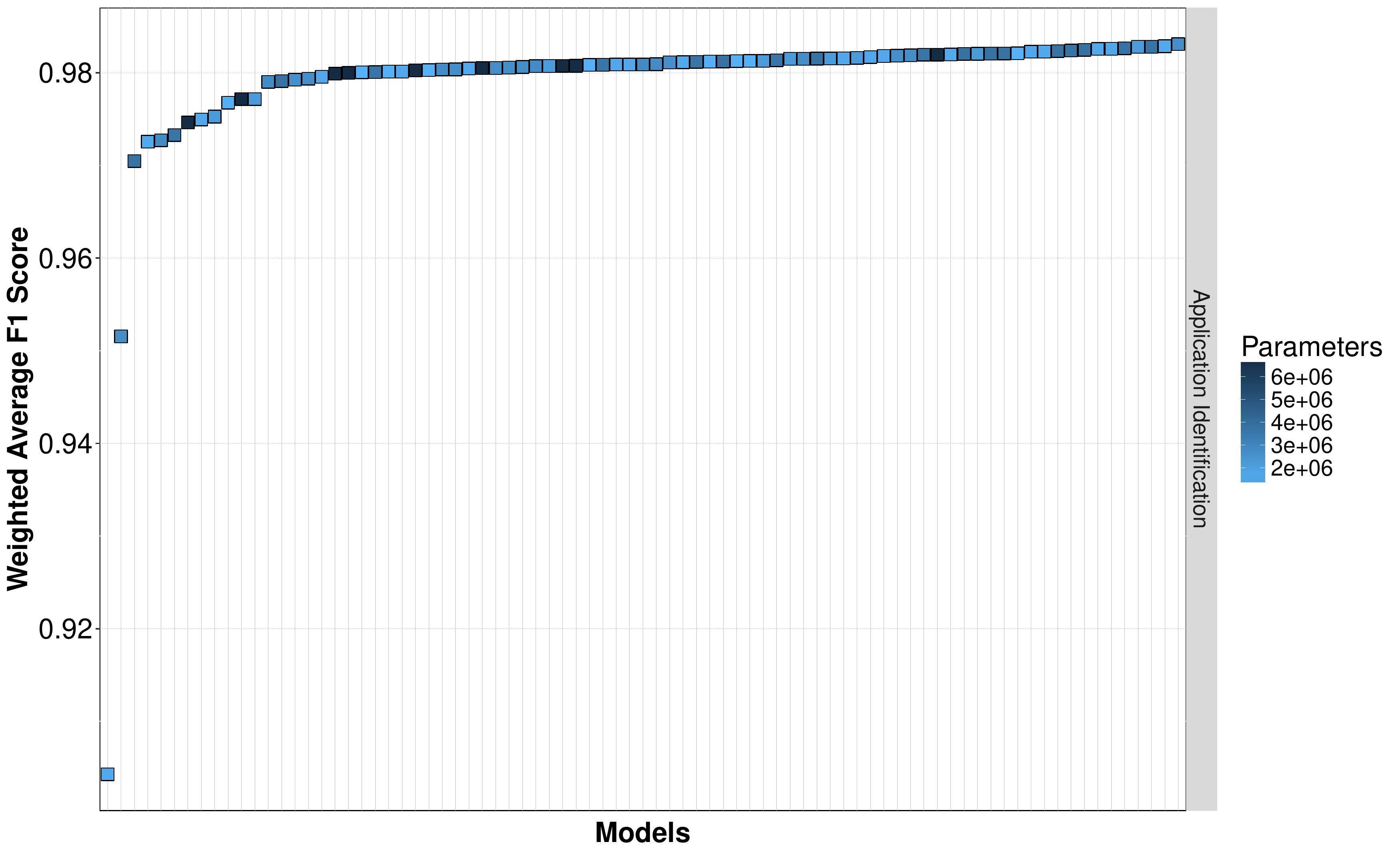}\label{figs:grd_idn}}
\par
\subfloat[][Traffic Characterization]{\includegraphics[width=0.9\linewidth,page=1]{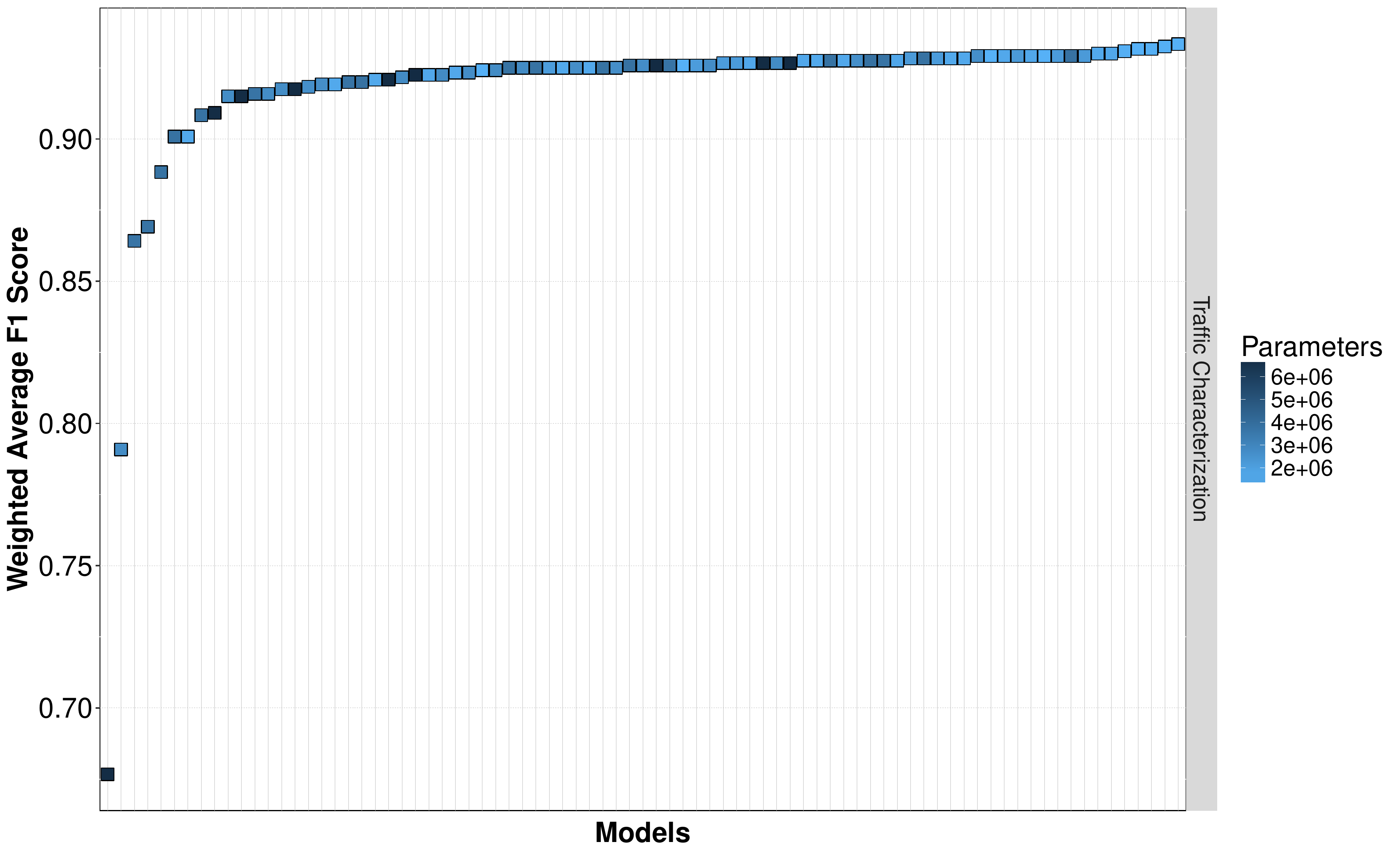}\label{figs:grd_char}}
\caption{Grid search on the hyperparameters of the proposed 1D-CNN for (a) Application identification, and (b) Traffic characterization.
}
\label{figs:grd}
\end{figure}

As mentioned in Section~\ref{sec:Methodology}, we used grid search hyper-parameters tuning scheme to find the best CNN structure. Due to our computation hardware limitations, we only searched a restricted subspace of hyper-parameters to find the ones which maximize the weighted average $F_1$ score on the test set for each task. To be more specific, we changed filter size, the number of filters and stride for both convolutional layers. In total, 116 models with their weighted average $F_1$ score for both application identification and traffic characterization tasks were evaluated. The result for all trained models, can be seen in Fig.~\ref{figs:grd}. We believe one cannot select an optimal model for traffic classification task since the definition of ``optimal model"  is not well-defined and there exists a trade-off between the model accuracy and its complexity (\ie, training and test speed). In Fig.~\ref{figs:grd}, the color of each point is associated with the model's trainable parameters; the darker the color, the higher the number of trainable parameters.

As seen in Fig.~\ref{figs:grd}, increasing the complexity of the neural network does not necessarily result in a better performance. 
Many reasons can cause this phenomenon which among them one can mention to the vanishing gradient and over-fitting problems. A complex model is more likely to face the vanishing gradient problem which leads to under-fitting in the training phase. On the other hand, if a learning model becomes more complex while the size of training data remains the same, the over-fitting problem can be occurred. Both of these problems lead to a poor performance of NNs in the evaluation phase.


%

Table~\ref{table:results-identification} shows the achieved performance of both SAE and CNN for the application identification task on the test set. The weighted average $F_1$ score of $0.98$ and $0.95$ for 1D-CNN and SAE, respectively, shows that our networks have entirely extracted and learned the discriminating features from the training set and can successfully distinguish each application. For the traffic characterization task, our proposed CNN and SAE have achieved $F_1$ score of $0.93$ and $0.92$ respectively, implying that both networks are capable of accurately classify packets. Table~\ref{table:results-charac} summaries the achieved performance of the proposed methods on the test set.
%
%
\begin{table}[!htb]
    \centering
    \scriptsize
    \caption[caption]{Application identification results.}
    \label{table:results-identification}
    \renewcommand{\arraystretch}{1.2}
	\begin{tabular}{| c | c  c  c |  c  c  c |}
	\hline
	 \multirow{2}{*}{Application} & \multicolumn{3}{c|}{ CNN }  &  \multicolumn{3}{c|}{ SAE }\\
	\cline{2-7}
	 & Rc & Pr & $F_1$  & Rc & Pr & $F_1$ \\
    \hline
AIM chat & 0.76 & 0.87 & 0.81 & 0.64 & 0.76 & 0.70 \\ 
Email & 0.82 & 0.97 & 0.89 & 0.99 & 0.94 & 0.97\\

Facebook & 0.95 & 0.96 & 0.96 & 0.95 & 0.94 & 0.95 \\

FTPS & 1.00 & 1.00 & 1.00 & 0.77 & 0.97 & 0.86 \\ 

Gmail &  0.95 & 0.97 & 0.96 & 0.94 & 0.93 & 0.94 \\

Hangouts & 0.98 & 0.96 & 0.97 & 0.99 & 0.94 & 0.97 \\ 

ICQ & 0.80 & 0.72 & 0.76 & 0.69 & 0.69 & 0.69 \\ 
Netflix & 1.00 & 1.00 & 1.00 & 1.00 & 0.98 & 0.99 \\
SCP & 0.99 & 0.97 & 0.98 & 1.00 & 1.00 & 1.00 \\ 
SFTP & 1.00 & 1.00 & 1.00 & 0.96 & 0.70 & 0.81 \\ 
Skype & 0.99 & 0.94 & 0.97 & 0.93 & 0.95 & 0.94 \\ 
Spotify & 0.98 & 0.98 & 0.98 & 0.98 & 0.98 & 0.98 \\ 
Torrent & 1.00 & 1.00 & 1.00 & 0.99 & 0.99 & 0.99 \\
Tor & 1.00 & 1.00 & 1.00 & 1.00 & 1.00 & 1.00 \\ 

VoipBuster & 1.00 & 0.99 & 0.99 & 0.99 & 0.99 & 0.99 \\ 
Vimeo & 0.99 & 0.99 & 0.99 & 0.98 & 0.99 & 0.98 \\ 
YouTube & 0.99 & 0.99 & 0.99 & 0.98 & 0.99 & 0.99 \\
\hline
\hline
\textbf{Wtd. Average} & \textbf{0.98} & \textbf{0.98} & \textbf{0.98} & \textbf{0.96} & \textbf{0.95} & \textbf{0.95} \\
\hline
	\end{tabular}
\end{table}

\begin{table}[!ht]
    \centering
    \caption[caption]{Traffic characterization results.}
	\hspace*{-0.75cm}

\label{table:results-charac}
	\renewcommand{\arraystretch}{1.2}
	\scriptsize
	\begin{tabular}{|c | c  c  c  | c  c  c |}
	\hline
	\multirow{2}{*}{Class Name} & \multicolumn{3}{c|}{ CNN }  &  \multicolumn{3}{c|}{ SAE }\\
	\cline{2-7}
	 & Rc & Pr & $F_1$  & Rc & Pr & $F_1$ \\ 
	\hline
	Chat& 0.71 & 0.84 & 0.77 & 0.68 & 0.82 & 0.74  \\ 
	Email& 0.87 & 0.96 & 0.91 & 0.93 & 0.97 & 0.95 \\ 
	File Transfer & 1.00 & 0.98 & 0.99 & 0.99 & 0.98 & 0.99 \\ 
	Streaming & 0.87 & 0.92 & 0.90 & 0.84 & 0.82 & 0.83 \\ 
	Torrent& 1.00 & 1.00 & 1.00 & 0.99 & 0.97 & 0.98\\ 
	VoIP& 0.88 & 0.63 & 0.74 & 0.90& 0.64 & 0.75 \\ 
	VPN: Chat & 0.98 & 0.98 & 0.98 & 0.94 & 0.95 & 0.94\\ 
	VPN: File Transfer & 0.99 & 0.99 & 0.99 & 0.95 & 0.98 & 0.97 \\ 
	VPN: Email & 0.98 & 0.99 & 0.99 & 0.93 & 0.97 & 0.95\\ 
 	VPN: Streaming & 1.00 & 1.00 & 1.00 & 0.99 & 0.99 & 0.99 \\ 
	VPN: Torrent & 1.00 & 1.00 & 1.00 & 0.97 & 0.99 & 0.98  \\ 
	VPN: VoIP & 1.00 & 0.99 & 1.00 & 1.00 & 0.99 & 0.99  \\ 
	\hline
	\hline
	\textbf{Wtd. Average} & \textbf{0.94}  & \textbf{0.93} & \textbf{0.93} & \textbf{0.92} & \textbf{0.92} & \textbf{0.92}\\
	\hline
\end{tabular}

\end{table}

As mentioned in Section~\ref{sec:relatedWorks}, authors in \cite{gil2016characterization} tried to characterize network traffic using time-related features handcrafted from traffic flows such as the duration of the flow and flow bytes per second. Yamansavascilar \etal~also used such time-related features to identify the end-user application \cite{yamansavascilar2017application}. Both of these studies evaluated their models on the ``ISCX VPN-nonVPN traffic dataset'' and their best results can be found in Table~\ref{table:comp-result}. The results suggest that Deep Packet has outperformed other proposed approaches mentioned above, in both application identification and traffic characterization tasks.

We would like to emphasize that the above-mentioned work have used hand-crafted features based on the network traffic flow. On the other hand, Deep Packet considers the network traffic in the packet-level and can classify each packet of network traffic flow which is a harder task, since there are more information in a flow compared to a single packet. This feature allows Deep Packet to be more applicable in real-world situations.

Finally, it worth mentioning that independently and parallel to our work \cite{DeepPacketArXiv}, Wang \etal~proposed a similar approach to Deep Packet for traffic characterization on ``ISCX VPN-nonVPN" traffic dataset \cite{wang_end--end_2017}. Their best-reported result achieves 100\% precision on the traffic characterization task. However, we believe that their result is seriously questionable. The proving reason for our allegation is that their best result has been obtained by using packets containing all the headers from every five layers of  the Internet protocol stack. However based on our experiments and also a direct inquiry from the dataset providers \cite{gil2016characterization}, in ``ISCX VPN-nonVPN'' traffic dataset, the source and destination IP addresses (that are appeared in the header of network layer) are unique for each application.
Therefore, their model presumably just uses this feature to classify the traffic (in that case a much simpler classifier would be sufficient to handle the classification task). As mentioned before, to avoid this phenomenon, we mask IP address fields in the preprocessing phase before feeding the packets into our NNs for training or testing.

\begin{table}[!htb]
\caption{A comparison between Deep Packet and other proposed methods on ``ISCX VPN-nonVPN" dataset.}
\label{table:comp-result}
\renewcommand{\arraystretch}{1.2}
\centering
\scriptsize
\makebox[\linewidth]{
\begin{tabular}{|c|c|c|c|c|}

\hline
Paper                  & Task                       &  Metric & Results       & Algorithm   \\
\hline
Deep Packet   & Application  & 	\multirow{2}{*}{Accuracy}    & 0.98   & CNN          \\ 
\cline{4-5}
\cline{0-0}
\cite{yamansavascilar2017application} & Identification &   & 0.94        & k-NN           \\
\hline
Deep Packet   & Traffic     & \multirow{2}{*}{Precision}     & 0.93   & CNN   \\ 

\cline{4-5}
\cline{0-0}

\cite{gil2016characterization}   &  Characterization  &   & 0.90 &  C4.5      \\ 
\hline
\end{tabular}}
\end{table}

\section{Discussion}\label{sec:discus}
\begin{figure}[!htb]
\centering
\subfloat[][]{\includegraphics[width=0.47\linewidth,page=2]{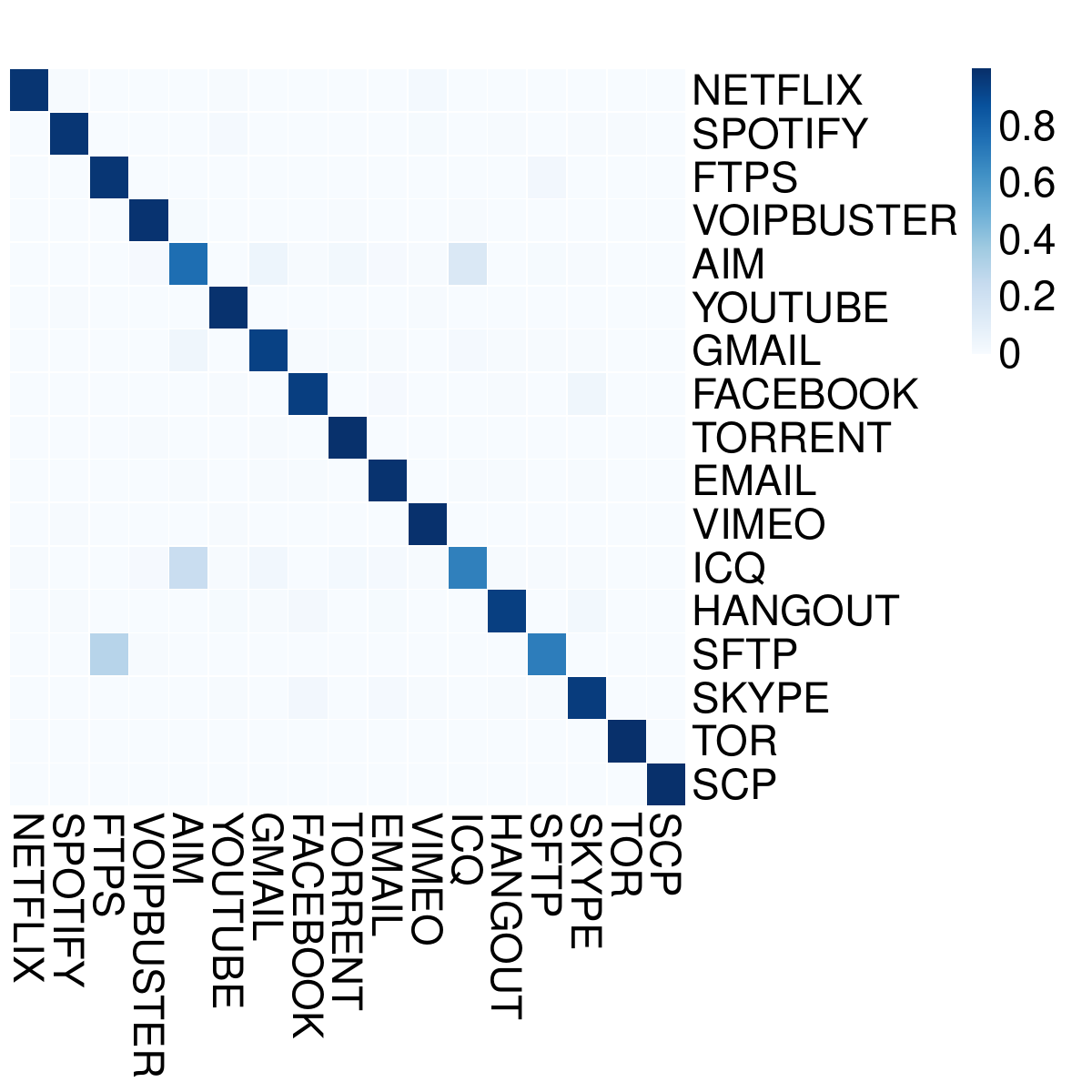}\label{figs:heatmap-idn-sae}}
\subfloat[][]{\includegraphics[width=0.47\linewidth,page=2]{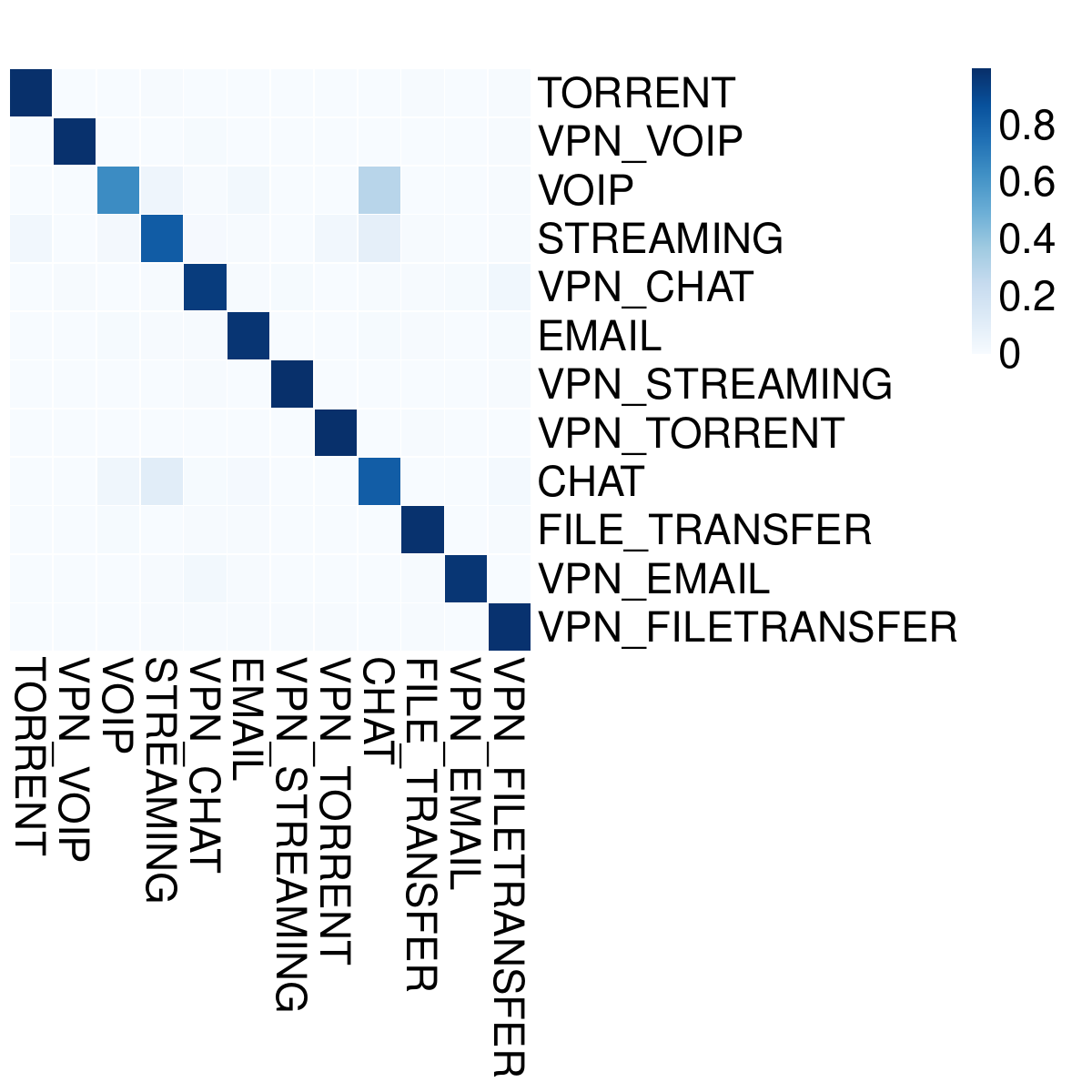}\label{figs:heatmap-char-sae}}
\caption{Row-normalized confusion matrices using SAE on (a) Application identification, and (b) Traffic characterization.
}
%
\label{figs:heatmaps}
\end{figure}

Evaluating the SAE on the test set for the Application Identification and the Traffic Characterization tasks result in row-normalized confusion matrices shown in Fig.~\ref{figs:heatmaps}.
 The rows of the confusion matrices correspond to the actual class of the samples, and the columns present the predicted label, thus the matrices are row-normalized. The dark color of the elements on the main diagonal suggests that SAE can classify each application with minor confusion.
%

\begin{figure}[!htb]
\centering
\subfloat[][Application identification \\using SAE.]{\includegraphics[width=0.49\linewidth,page=1]{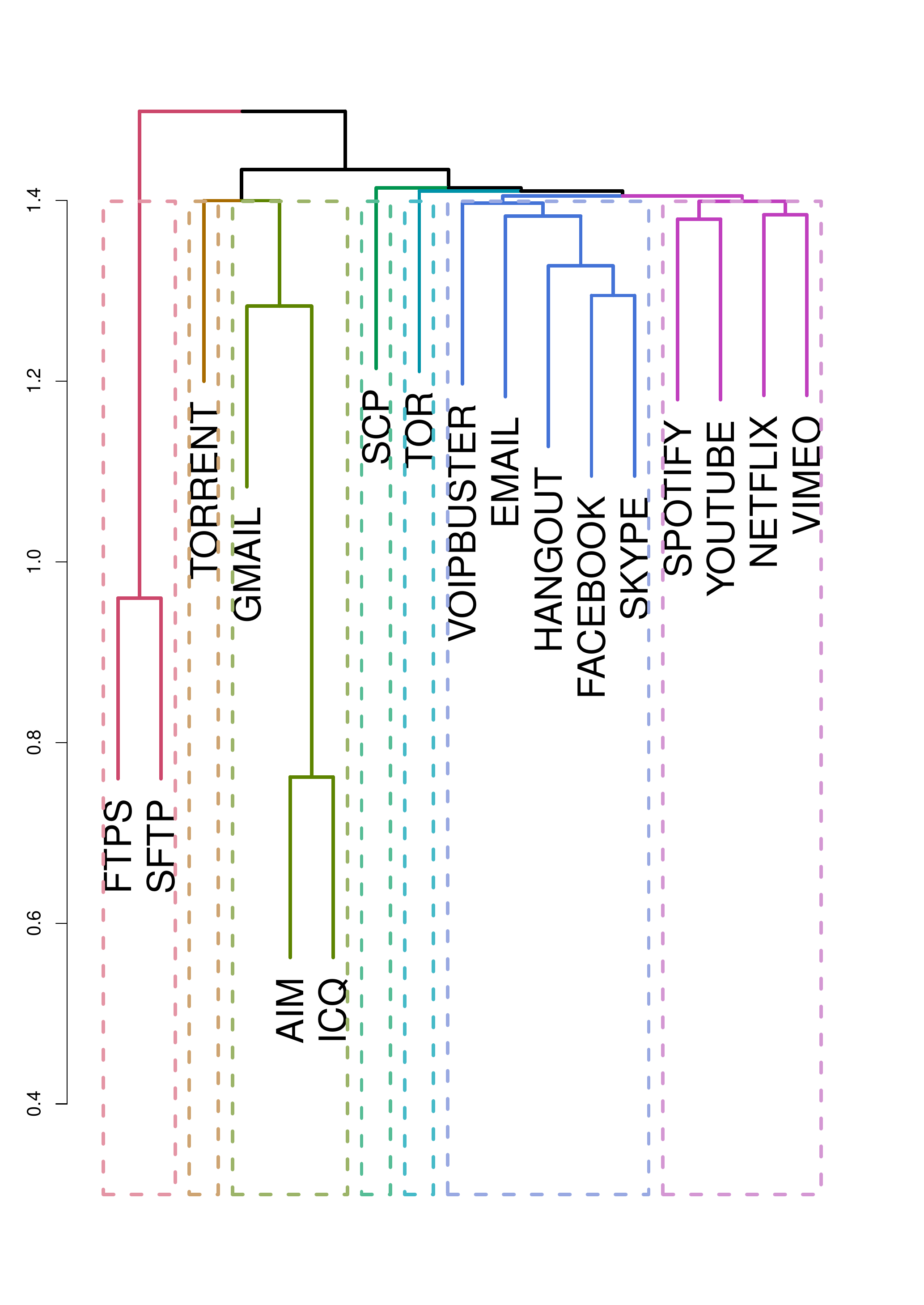}\label{figs:cls-idn-sae}}
\subfloat[][Traffic characterization \\using SAE.]{\includegraphics[width=0.49\linewidth,page=1]{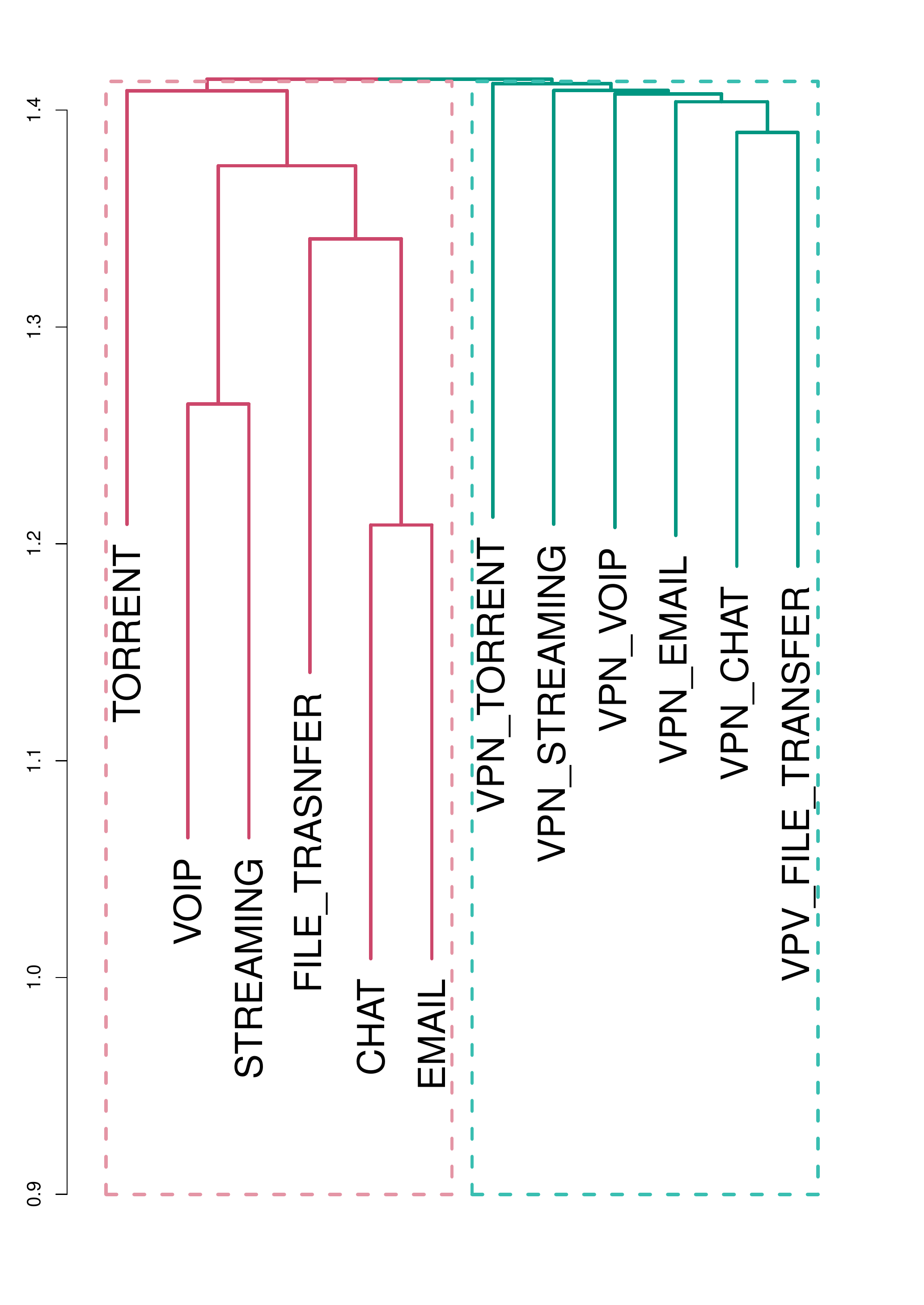}\label{figs:cls-char-sae}}
\caption{Hierarchical clustering, performed on row-normalized confusion matrices of the proposed SAE network.
}%
\label{figs:clusters}
\end{figure}

By carefully observing the confusion matrices in Fig.~\ref{figs:heatmaps}, one would notice some interesting confusion between different classes (\eg, ICQ, and AIM).
\emph{Hierarchical clustering} further demonstrates the similarities captured by Deep Packet. Clustering on row-normalized confusion matrices for application identification with SAE (Fig.~\ref{figs:heatmap-idn-sae}), using \emph{Euclidean distance} as the distance metric and \emph{Ward.D} as the agglomeration method uncovers similarities among applications regarding their propensities to be assigned to the $17$ application classes. As illustrated in Fig.~\ref{figs:cls-idn-sae}, application groupings revealed by Deep Packet, generally agrees with the applications’ similarities in the real world. Hierarchical clustering divided the applications into $7$ groups. Interestingly, these groups are to some extent similar to groups in the traffic characterization task. One would notice that Vimeo, Netflix, YouTube and Spotify which are bundled together, are all streaming applications. There is also a cluster including ICQ, AIM, and Gmail. AIM and ICQ are used for online chatting and Gmail in addition to email services, offers a service for online chatting. Another interesting observation is that Skype, Facebook, and Hangouts are all grouped in a cluster together. Though these applications do not seem much relevant, this grouping can be justified. The dataset contains traffic for these applications in three forms: voice call, video call, and chat. Thus the network has found these applications similar regarding their usage. FTPS (File Transfer Protocol over SSL) and SFTP (File Transfer Protocol over SSH) which are both used for transferring files between two remote systems securely are clustered together as well. Interestingly, SCP (Secure Copy) has formed its cluster although it is also used for remote file transfer. SCP uses SSH protocol for transferring file, while SFTP and FTPS use FTP. Presumably, our network has learned this subtle difference and separated them. Tor and Torrent have their clusters which are sensible due to their apparent differences with other applications. This clustering is not flawless. Clustering Skype, Facebook, and Hangouts along with Email and VoipBuster are not correct.  VoipBuster is an application which offers voice communications over Internet infrastructure. Thus applications in this cluster do not seem much similar regarding their usage, and this grouping is not precise. 

The same procedure was performed on the confusion matrices of traffic characterization as illustrated in Fig.~\ref{figs:cls-char-sae}. Interestingly, groupings separates the traffic into VPN and non-VPN clusters. All the VPN traffics are bundled together in one cluster, while all of non-VPNs are grouped together.  




As mentioned in Section~\ref{sec:relatedWorks}, many of the applications employ encryption to maintain clients' privacy. As a result, the majority of ``ISCX VPN-nonVPN" dataset traffics are also encrypted. One might wonder how it is possible for Deep Packet to classify such encrypted traffics. Unlike DPI methods, Deep Packet does not inspect the packets for keywords. In contrast, it attempts to learn features in traffic generated by each application. Consequently, it does not need to decrypt the packets to classify them.

An ideal encryption scheme causes the output message to bear the maximum possible entropy \cite{Cover:2006:EIT:1146355}. In other words, it produces patternless data that theoretically cannot be distinguished from one another. However, due to the fact that all practical encryption schemes use pseudo-random generators, this hypothesis is not valid in practice. Moreover, each application employs different (non-ideal) ciphering scheme for data encryption. These schemes utilize different pseudo-random generator algorithms which leads to distinguishable patterns. Such variations in the pattern can be used to separate applications from one another. Deep Packet attempts to extract those discriminative patterns and learns them. Hence, it can classify encrypted traffic accurately.

It is noticeable from Table~\ref{table:results-identification} that Tor traffic is also successfully classified. To further investigate this kind of traffic, we conducted another experiment in which we trained and tested Deep Packet with a dataset containing only Tor traffic. To achieve the best possible result, we performed a grid search on the hyper-parameters of the NN, as discussed before. The detailed results can be found in Table~\ref{table:results-tor}, which shows that Deep Packet was unable to classify the underlying Tor's traffic accurately. This phenomenon is not far from what we expected. Tor encrypts its traffic, before transmission. As mentioned earlier, Deep Packet presumably learns pseudo-random patterns used in encryption scheme used by application. At this experiment, traffic was tunneled through Tor. Hence they all experience the same encryption scheme. Consequently, our neural network was not able to separate them apart. 

\begin{table}[!htb]
    \centering
    \scriptsize
    \caption[caption]{Tor traffic classification results. }
\label{table:results-tor}
	\renewcommand{\arraystretch}{1.2}
	\begin{tabular}{|c | c  c  c  | c  c  c |}
	\hline
	\multirow{2}{*}{Class Name} & \multicolumn{3}{c|}{ CNN }  &  \multicolumn{3}{c|}{ SAE }\\
	\cline{2-7}
	 & Rc & Pr & $F_1$  & Rc & Pr & $F_1$ \\ 
	\hline
	
	Tor: Google & 0.00 & 0.00 & 0.00 & 0.44 & 0.03 & 0.06 \\
		
	Tor: Facebook& 0.24&0.10&0.14 & 0.28 & 0.06 & 0.09  \\ 
	Tor: YouTube& 0.44&0.55&0.49  & 0.44 &	 0.99 & 0.61 \\ 
	Tor: Twitter & 0.17&0.01&0.01 & 0.37 & 0.00 & 0.00 \\ 
	Tor: Vimeo & 0.36 & 0.44 & 0.40 & 0.91 & 0.05 & 0.09 \\
	\hline
	\hline
	\textbf{Wtd. Average} & \textbf{0.35} & \textbf{0.40} & \textbf{0.36} & \textbf{0.57} & \textbf{0.44} & \textbf{0.30} \\
	\hline
\end{tabular}
\end{table}

\vspace{-0.5cm}
\section{Conclusion}\label{sec:conc}
 In this paper, we presented Deep Packet, a framework that automatically extracts features from network traffic using deep learning algorithms to classify traffic. To the best of our knowledge, Deep Packet is the first traffic classification system using deep learning algorithms, namely, SAE and one-dimensional CNN that can handle both application identification and traffic characterization tasks. Our results showed that Deep Packet outperforms all of the similar works on the ``ISCX VPN-nonVPN" traffic dataset,  both in application identification and traffic characterization to the date. Moreover, with state-of-the-art results achieved by Deep Packet, we envisage that Deep Packet is the first step toward a  general trend of using deep learning algorithms in traffic classification tasks. Furthermore, Deep Packet can be modified to handle more complex tasks like multi-channel (e.g., distinguishing between different types of Skype traffic including chat, voice call, and video call) classification, accurate classification of Tor's traffic, etc. Finally, the automatic feature extraction procedure from network traffic can save the cost of using experts to identify and extract handcrafted features from the traffic which eventually leads to more accurate traffic classification.

\begin{acknowledgements}
The authors would like to thank Farzad Ghasemi and Mehdi Kharrazi for their valuable discussions and feedback.
\end{acknowledgements}

\section*{Compliance with Ethical Standards}
Conflict of Interest: Mohammad Lotfollahi, Mahdi Jafari Siavoshani, Ramin Shirali Hossein Zade, and Mohammdsadegh Saberian declare that they have no conflict of interest.\\
Ethical approval: This article does not contain any studies with human participants or animals performed by any of the authors.



\bibliographystyle{spbasic}      

\bibliography{ref}

\end{document}